\documentclass{article}
\usepackage{spconf,amsmath,graphicx}

\usepackage{enumitem}
\usepackage{graphicx} 
\usepackage{pifont}     
\usepackage{mwe} % to get dummy images
\usepackage{tikz}
\usepackage{threeparttable}
\usepackage{float}

\usepackage{tabularx} 

\usepackage{makecell}

\usepackage{graphicx}
\usepackage{amsmath}
\usepackage{booktabs}
\usepackage{multirow}
\usepackage{tikz}
\usetikzlibrary{shapes.geometric, arrows.meta, positioning, fit, backgrounds, calc}
\usepackage{pgfplots}
\pgfplotsset{compat=1.17}
\usepackage{xcolor}
\usepackage[draft,bookmarks=false]{hyperref}

\usepackage[T1]{fontenc} % Hỗ trợ font chữ tốt hơn
\usepackage{textcomp} % Cho các ký hiệu

\newcommand{\cmark}{\ding{51}}
\newcommand{\xmark}{\ding{55}} % Để tuỳ chỉnh list
\usepackage[font=small,skip=4pt]{caption} % Làm cho chú thích nhỏ hơn và gần hơn

\title{VinDr-CXR-VQA: A Visual Question Answering Dataset for Explainable Chest X-Ray Analysis with Multi-Task Learning}

\name{
    Dang H. Nguyen$^{1,2,4}$ \qquad 
    Hieu H. Pham$^{2,4}$ \qquad 
    Hao T. Nguyen$^{3}$ \qquad
    Hieu H. Pham$^{1,2,4*}$\thanks{*Correspondence: Huy-Hieu Pham (hieu.ph@vinuni.edu.vn)}
}

\address{
    $^{1}$College of Engineering and Computer Science, VinUniversity, 100000 Hanoi, Vietnam  \\
    $^{2}$VinUni-Illinois Smart Health Center, VinUniversity, 100000 Hanoi, Vietnam \\
    $^{3}$Quan Su Radiology Department, Vietnam National Cancer Hospital, 100000 Hanoi, Vietnam \\
    $^{4}$The Computer Vision and Medical AI Lab, VinUniversity, 100000 Hanoi, Vietnam \\
}

\begin{document}
%\ninept
%
\maketitle

\begin{abstract}
We present \textbf{VinDr-CXR-VQA}, a large-scale chest X-ray dataset for explainable Medical Visual Question Answering (Med-VQA) with spatial grounding. The dataset contains $17{,}597$ question-answer pairs across $4{,}394$ images, each annotated with radiologist-verified bounding boxes and clinical reasoning explanations. Our question taxonomy spans six diagnostic types: Where, What, Is there, How many, Which, and Yes/No, capturing diverse clinical intents. To improve reliability, we construct a balanced distribution of $41.7\%$ positive and $58.3\%$ negative samples, mitigating hallucinations in normal cases. Benchmarking with \textit{MedGemma-4B-it}, a state-of-the-art medical VLM, demonstrates improved performance (F1= $0.624$, $+11.8\%$ over baseline) while enabling lesion localization. VinDr-CXR-VQA aims to advance reproducible and clinically grounded Med-VQA research. The dataset and evaluation tools are publicly available at \textcolor{red}{\url{huggingface.co/datasets/Dangindev/VinDR-CXR-VQA}}.
\end{abstract}

\begin{figure}[t]
  \centering
  \includegraphics[width=.73\columnwidth]{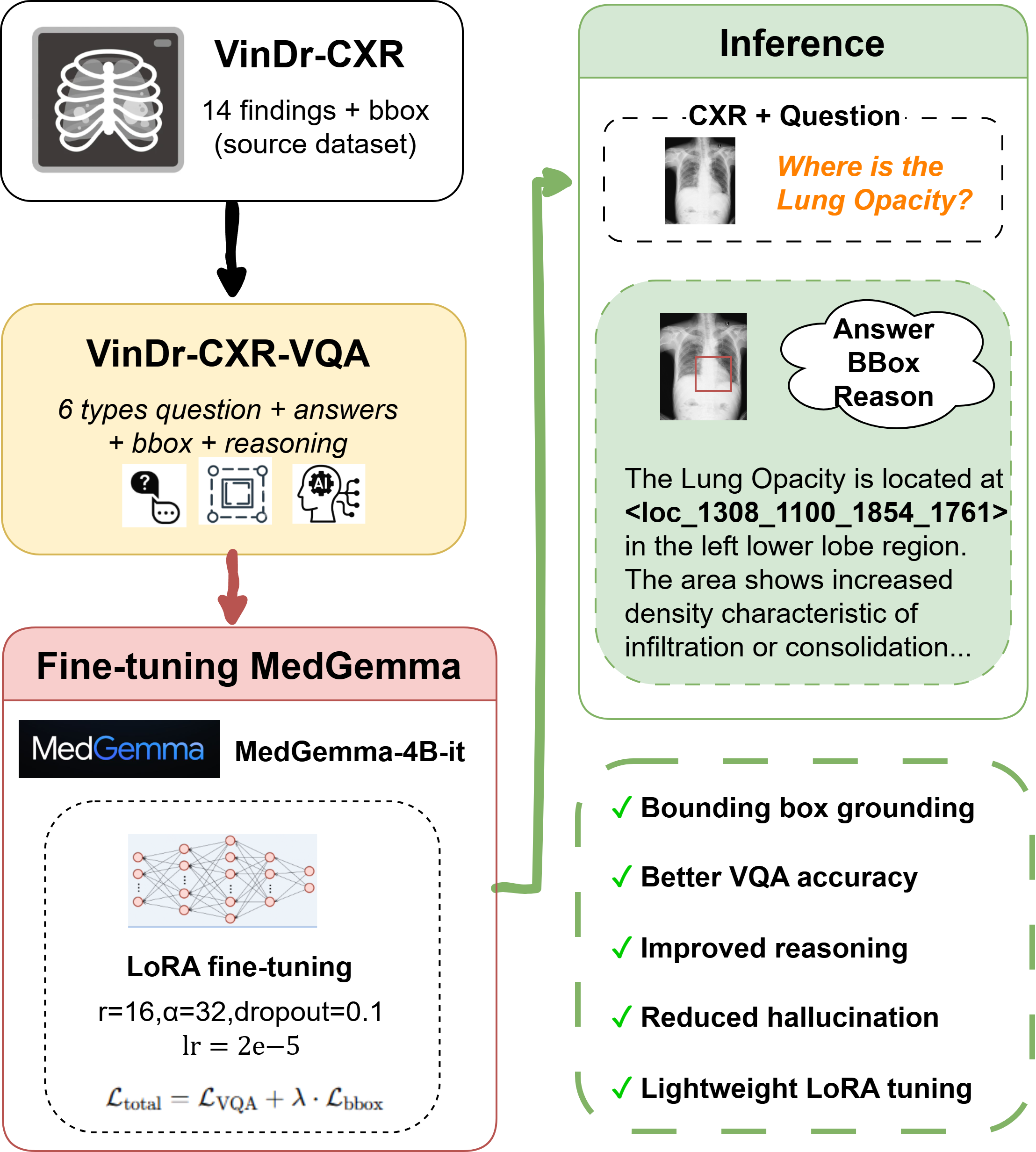}
  \caption{\textbf{Overview of the VinDr-CXR-VQA pipeline.} The figure illustrates the multi-task learning approach, encompassing dataset creation (unifying Question \& Answer, Bounding Box, and Reasoning) and the downstream fine-tuning of MedGemma-4B-it, enabling explainable VQA with accurate spatial grounding.}
  \label{fig:overview}
\end{figure}

\begin{table*}[t]
\centering
\begin{threeparttable}
\resizebox{0.85\textwidth}{!}{%
\begin{tabular}{lccccc}
\hline
\textbf{Dataset} & \textbf{Num. of Images} & \textbf{Num. of QA Pairs} & \textbf{Modalities} & \textbf{Groundable} & \textbf{Explainable} \\
\hline
VQA-RAD~\cite{vqa-rad} & 0.3K & 3.5K & Diverse\textsuperscript{\dag} & \xmark & \xmark \\
SLAKE~\cite{slake} & 0.6K & 14K & Diverse\textsuperscript{\dag} & \xmark & \xmark \\
PathVQA~\cite{pathvqa} & 149K & 33K & Pathology & \xmark & \xmark \\
MIMIC-CXR-VQA~\cite{mimic-cxr-vqa} & 143K & 377K & Chest X-ray & \xmark & \xmark \\
\hline
\textbf{Ours (VinDr-CXR-VQA)} & \textbf{4.4K} & \textbf{17.6K} & \textbf{Chest X-ray} & \textbf{\cmark} & \textbf{\cmark} \\
\hline
\end{tabular}
} % end resizebox
\begin{tablenotes}
\item[\dag] Diverse: multiple body parts and imaging modalities.
\item[\ddag] Groundable: provides spatial annotations (e.g., bounding boxes or heatmaps) tied to the answer.
\item[$\S$] Explainable: includes textual clinical reasoning or justification alongside the answer.
\end{tablenotes}
\caption{Comparison of medical VQA datasets. VinDr-CXR-VQA is the only dataset that provides both spatial grounding (bounding boxes) and explainability (clinical reasoning).}
\label{tab:dataset_comparison}
\end{threeparttable}
\end{table*}

\section{Introduction and Related Work}

Medical Visual Question Answering (Med-VQA) aims to answer clinical questions from medical images by combining visual recognition with natural language reasoning~\cite{med_vqa_survey}. For clinical applicability, it is crucial that models not only generate diagnostic answers but also provide spatial justifications~\cite{chest-imagenome,gemex}. Radiologists must be able to assess both \textit{what} the model detects and \textit{where} it localizes findings in order to validate AI outputs and ensure patient safety. Recent medical Vision-Language Models (VLMs), such as MedGemma~\cite{medgemma} and LLaVA-med~\cite{llava-med}, show promise but still require explicit spatial grounding for clinical validation. Beyond these, recent efforts in medical Large Vision-Language Models (LVLMs) and benchmarks further expand capabilities and evaluation breadth, including PaliGemma~\cite{paligemma}, RadFM~\cite{radfm}, and XrayGPT~\cite{xraygpt}.

However, existing Med-VQA datasets fall short. Text-only benchmarks lack spatial annotations, while detection datasets (VinDr-CXR~\cite{vindr-cxr}, ChestX-ray14~\cite{chestxray14}) miss interactive Question Answering (QA) capabilities~\cite{vindr-cxr}. Even broader grounded efforts still lack precise VQA grounding for chest X-rays. To the best of our understanding, this critical gap necessitates a new dataset unifying clinical QA and spatial localization.

To this end, we present \textbf{VinDr-CXR-VQA}, a grounded Med-VQA dataset of $4{,}394$ images and $17{,}597$ QA pairs, verified by radiologists. The dataset spans six diagnostic types and includes bounding boxes and clinical reasoning. We ensured robust training by balancing positive ($41.7\%$) and negative ($58.3\%$) samples. VinDr-CXR-VQA is uniquely positioned to offer both spatial grounding and clinical explainability (Table~\ref{tab:dataset_comparison} and Figure~\ref{fig:overview}).

\noindent\textbf{Our contributions are:}
\begin{itemize}[leftmargin=*,noitemsep,topsep=2pt]
    \item We introduce \textbf{VinDr-CXR-VQA}, a chest X-ray dataset that combines spatial grounding (bounding boxes) with clinical question-answering and reasoning explanations.
    \item The dataset contains 17.6K QA pairs across 4.4K radiographs, covering six structured question types and a balanced distribution of normal and abnormal findings.
    \item All annotations, including bounding boxes and textual explanations, are curated and verified by board-certified radiologists.
    \item The dataset is publicly released to support the development and evaluation of explainable, clinically grounded Med-VQA models.
\end{itemize}

The remainder of this paper details the dataset construction, followed by our experimental setup and evaluation, and concludes with a discussion of future directions.

\section{VinDr-CXR-VQA Dataset}

\subsection{Dataset Construction}

VinDr-CXR~\cite{vindr-cxr} provides 18{,}000 posteroanterior chest X-ray images with expert-validated bounding box annotations across 22 thoracic diseases, establishing a gold standard for chest X-ray object detection. To reduce annotation noise and ensure sufficient training samples per class, we select 14 clinically relevant pathologies with higher prevalence, excluding rare or ambiguous findings. However, VinDr-CXR lacks the natural language supervision required for vision-language model training. To address this, we build VinDr-CXR-VQA by adding question-answer pairs and clinical reasoning, while preserving all expert annotations.

We selected 4,394 images from VinDr-CXR containing at least one expert-annotated pathology finding. The authors manually designed structured prompt templates for six question categories: \textit{Where} (spatial localization), \textit{What} (pathology identification), \textit{Is\_there} (existence verification), \textit{How\_many} (counting), \textit{Which} (anatomical classification), and \textit{Yes\_No} (binary confirmation). Each template directly references verified VinDr-CXR pathology labels and bounding boxes, grounding generated questions in validated clinical findings.

We utilize Google's Gemini 2.5 Pro vision-language API~\cite{gemini_2_5_2025} to systematically generate the required natural language content. For each image-template combination, the API receives the chest X-ray, along with its full VinDr-CXR annotation (which includes the pathology label and bounding box), as input. The API then generates three distinct components: (1) a coherent natural language question, (2) a precise answer incorporating spatial reference in \texttt{<loc\_xmin\_ymin\_xmax\_ymax>} format, and (3) a detailed clinical reasoning paragraph (150-250 words). This reasoning, which constitutes the core medical knowledge component and offers crucial explainability, is entirely generated by Gemini 2.5 Pro based on its extensive medical literature training data. The API articulates diagnostic significance, differential diagnoses, and clinical implications using appropriate medical terminology. Critically, the API only generates textual descriptions – all pathology labels and bounding box coordinates are directly copied from VinDr-CXR without modification, ensuring evaluation metrics assess model performance against verified clinical expertise.

\begin{table}[h]
\centering

\resizebox{0.95\columnwidth}{!}{%
\begin{tabular}{llp{4.5cm}}
\toprule
\textbf{Field} & \textbf{Source} & \textbf{Description} \\
\midrule
\texttt{question} & API & Natural language query \\
\texttt{answer} & API & Response with \texttt{<loc>} reference \\
\texttt{reason} & API & Clinical reasoning (150-250 words) \\
\texttt{type} & API & Question category (6 types) \\
\texttt{difficulty} & API & Easy/Medium \\
\midrule
\texttt{gt\_finding} & VinDr-CXR & Expert pathology label (unchanged) \\
\texttt{gt\_location} & VinDr-CXR & Expert bounding box (unchanged) \\
\bottomrule
\end{tabular}
}
\caption{JSON Schema with Source Attribution}
\label{tab:json_structure}

\end{table}

Table~\ref{tab:json_structure} presents the dataset's JSON schema. Each sample contains seven fields: five generated by the API (question, answer, reason, type, difficulty) and two directly inherited from VinDr-CXR (gt\_finding, gt\_location). This clear separation between generated content and preserved ground truth is fundamental to the dataset's reliability.

\subsection{Quality Control and Clinical Validation}

To ensure dataset reliability, we performed rigorous quality control via automated verification and clinical expert review.

\noindent\textbf{Automated Verification.} Automated Verification involves programmatic validation, ensuring 100\% preservation of expert annotations. Scripts meticulously verified structural integrity across all seven JSON fields. Systematic cross-validation of all 4,394 images further confirmed zero mismatches against VinDr-CXR source files, ensuring 100\% preservation of expert annotations.

\noindent\textbf{Clinical Expert Review.} Clinical Expert Review involved independent evaluation of 100 question-answer pairs (0.57\% sample) by two board-certified radiologists (8 years experience). Reviewers assessed three dimensions (accuracy of reasoning, answer appropriateness, and spatial reference correctness). Inter-rater agreement was near-perfect ($\kappa = 0.89$ [95\% CI: 0.84, 0.93]). Initial disagreements (7 cases) were resolved via consensus, resulting in all 100 samples receiving "acceptable" ratings across all dimensions.

This multi-tier process provides strong evidence of trustworthiness, ensuring perfect ground truth preservation and medical accuracy via high inter-rater reliability.

\subsection{Dataset Statistics and Composition}

\begin{table}[h]
\centering

\resizebox{\columnwidth}{!}{%
\begin{tabular}{lc}
\toprule
\textbf{Attribute} & \textbf{Value} \\
\midrule
Source images (VinDr-CXR) & 4,394 \\
Generated Q\&A pairs & 17,597 \\
Average Q\&A per image & 4.0 \\
Pathology classes & 14 \\
Question types & 6 \\
\midrule
Training split & 3,735 images (85\%) \\
Validation split & 659 images (15\%) \\
Test split & 300 images \\
\midrule
Difficulty distribution & Easy: 49.1\%, Medium: 50.9\% \\
Question type distribution & Balanced (16.4–17.0\% each) \\
Multi-lesion complexity (validation) & 54.8\% with $\geq$2 pathologies \\
Average bboxes per validation image & 8.3 \\
\bottomrule
\end{tabular}
}
\caption{VinDr-CXR-VQA Dataset Statistics}
\label{tab:stats}

\end{table}

\noindent Table~\ref{tab:stats} summarizes the comprehensive dataset composition. From VinDr-CXR's 18,000 images, we generated 17,597 question-answer pairs from 4,394 images (avg 4.0 pairs/image), featuring balanced distribution across six question types (16.4-17.0\% each). The dataset spans 14 pathology classes with natural imbalance ranging from Cardiomegaly (28.0\%) to Atelectasis (0.1\%), accurately reflecting real-world clinical prevalence. This structure robustly enables assessment of model performance across both common and rare pathologies.

Data splits follow VinDr-CXR's official partitioning with stratified sampling to maintain class proportions. Validation and test sets preserve multi-lesion complexity: 54.8\% of validation images contain $\geq$2 pathologies with an average of 8.3 bounding boxes per image, reflecting realistic clinical scenarios where simultaneous multi-pathology detection and reasoning are required.

\begin{figure*}[t]
\centering
\includegraphics[width=\textwidth]{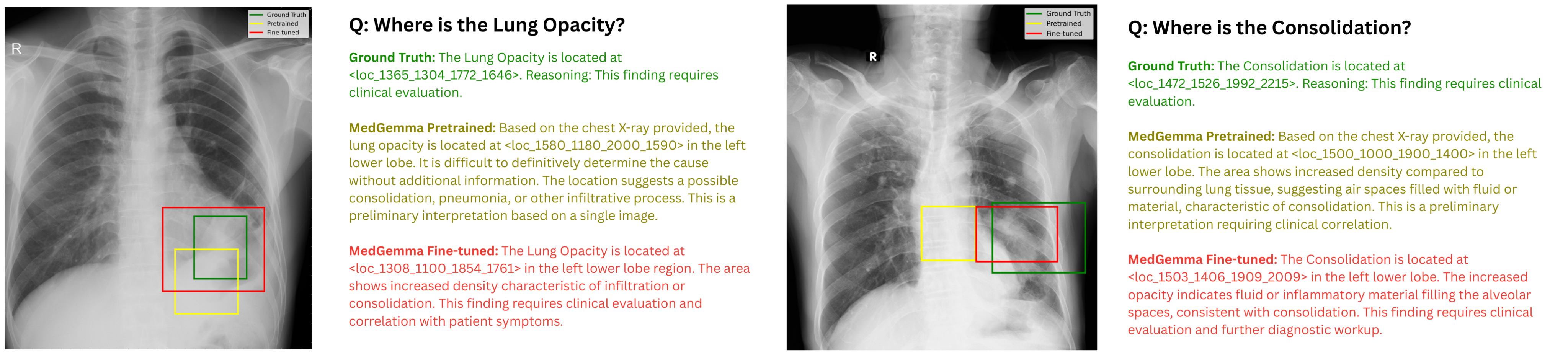}
\caption{Model comparison on VinDr-CXR validation images. Bounding boxes: ground truth (green), Predictions from MedGemma Pretrained (yellow), and Predictions from MedGemma Fine-tuned (red). (A) Left Image: Infiltration with high IoU (0.715). The Fine-tuned model achieves excellent localization, significantly outperforming the pretrained baseline. (B) Right Image: Consolidation with moderate IoU (0.482). The Fine-tuned model demonstrates improved accuracy over the pretrained baseline. Both cases show the fine-tuned model's superior spatial localization while maintaining clinical reasoning capability.}
\label{fig:qualitative}
\end{figure*}

\section{Experiments}
\subsection{Implementation Details}

We fine-tune MedGemma-4B-it~\cite{medgemma}, a 4B vision-language model pretrained on medical imaging, using Low-Rank Adaptation (LoRA)~\cite{lora} for parameter-efficient training. LoRA is applied with rank $r=16$, scaling factor $\alpha=32$, and dropout 0.1, adding only 0.21\% trainable parameters (8.9M parameters) while keeping the base model frozen. Optimization uses AdamW with learning rate $2 \times 10^{-5}$, weight decay 0.01, and batch size 4 with gradient accumulation steps of 4 (effective batch size 16). Training is conducted for 3 epochs over the 20,880-sample training set, requiring approximately 36 hours on a single NVIDIA A6000 GPU (48GB VRAM). Mixed precision training (fp16) is employed to reduce memory consumption. We save checkpoints at the end of each epoch and select the best model based on validation F1 score.

Our multi-task objective combines VQA and bounding box prediction: $\mathcal{L}_{total} = \mathcal{L}_{VQA} + \lambda \cdot \mathcal{L}_{bbox}$, with $\lambda = 0.5$. The VQA loss uses cross-entropy over answer tokens, while the bbox loss employs IoU-based regression. Bounding box coordinates are predicted via special tokens (\texttt{<loc\_x1\_y1\_x2\_y2>}) embedded in the generated answer, enabling unified end-to-end training without architectural modifications.

\subsection{Evaluation Metrics}

For \textbf{VQA performance}, we evaluate text-based lesion classification using Accuracy, Precision, Recall, and F1 score on the validation set (659 images, 5,471 ground-truth bounding boxes), comparing the pretrained baseline (zero-shot, no fine-tuning on our dataset) against our fine-tuned model.

For \textbf{bounding box localization}, we \textbf{specifically adopt standard} detection metrics with IoU threshold 0.3: a predicted box is a true positive (TP) if it overlaps any ground-truth box with IoU $\geq 0.3$, otherwise a false positive (FP). Missed ground-truth boxes are false negatives (FN). We report Precision, Recall, F1, and mean IoU (mIoU) computed over TPs only. The IoU threshold of 0.3 represents a clinically meaningful minimum overlap.

To assess \textbf{clinical localization quality} on a per-image basis, we measure the proportion of validation images achieving at least one prediction with IoU $\geq$ 0.5 ("good" localization) and IoU $\geq$ 0.3 ("acceptable" localization), plus mean IoU over all predictions.

\subsection{Quantitative Results}

\textbf{VQA Performance.} Table~\ref{tab:vqa} shows that fine-tuning on VinDr-CXR-VQA improves F1 from 0.558 to 0.624 (+11.8\%), with consistent gains across all metrics.

\begin{table}[h]
\centering
\small
\begin{tabularx}{\columnwidth}{l *{4}{>{\centering\arraybackslash}X}}
\hline
\textbf{Model} & \textbf{Acc.} & \textbf{Prec.} & \textbf{Recall} & \textbf{F1} \\
\hline
Pretrained & 0.558 & 0.421 & 0.827 & 0.558 \\
Fine-tuned & \textbf{0.624} & \textbf{0.471} & \textbf{0.926} & \textbf{0.624} \\
\hline
\end{tabularx}
\caption{VQA performance on 659 validation images.}
\label{tab:vqa}
\end{table}

\noindent\textbf{Bounding Box Detection.} Table~\ref{tab:bbox_detection} reports bbox detection performance. While recall remains low (0.090), the model achieves a high mIoU of 0.615 on true positives, confirming its ability to produce accurate spatial grounding when predictions are correct.

\begin{table}[h]
\centering
\small
\begin{tabularx}{\columnwidth}{l *{4}{>{\centering\arraybackslash}X}}
\hline
\textbf{Model} & \textbf{Prec.} & \textbf{Recall} & \textbf{F1} & \textbf{mIoU (TP)} \\
\hline
Pretrained & 0.096 & 0.068 & 0.070 & 0.036 \\
Fine-tuned & \textbf{0.120} & \textbf{0.090} & \textbf{0.103} & \textbf{0.615} \\
\hline
\end{tabularx}
\caption{Bounding box detection performance (IoU$\geq$0.3).}
\label{tab:bbox_detection}
\end{table}

\noindent\textbf{Localization Quality.} Table~\ref{tab:bbox_quality} summarizes spatial localization quality on a per-image basis. The fine-tuned model achieves "good" localization (IoU$\geq$0.5) in 22.8\% of cases, and "acceptable" localization (IoU$\geq$0.3) in 48.6\% of cases — substantially outperforming the pretrained baseline.

\begin{table}[h]
\centering
\small
\begin{tabularx}{\columnwidth}{l *{2}{>{\centering\arraybackslash}X}}
\hline
\textbf{Metric} & \textbf{Pretrained} & \textbf{Fine-tuned} \\
\hline
IoU$\geq$0.5 (\%) & 1.1 & \textbf{22.8} \\
IoU$\geq$0.3 (\%) & 8.5 & \textbf{48.6} \\
Mean IoU (all preds) & 0.036 & \textbf{0.133} \\
\hline
\end{tabularx}
\caption{Spatial localization quality (per image).}
\label{tab:bbox_quality}
\end{table}

\noindent\textbf{Multi-task Effect.} Fine-tuning improves the primary VQA task while simultaneously enabling spatial grounding, demonstrating the effectiveness of VinDr-CXR-VQA as a multi-modal, clinically grounded supervision source.

\subsection{Qualitative Results}

Figure~\ref{fig:qualitative} illustrates examples of improved localization after fine-tuning. Ground truth boxes are shown in green, predictions from the pretrained model in yellow, and predictions from the fine-tuned model in red. The examples highlight the model's ability to accurately localize abnormalities (e.g., infiltration, consolidation) while preserving clinical reasoning in its textual responses.

\section{Discussion and Conclusion}

VinDr-CXR-VQA presents a large-scale, publicly available dataset designed to support spatially-grounded medical VQA, consisting of 17,597 question-answer pairs over 4,394 chest X-rays. Each sample is annotated with expert-validated bounding boxes and clinical reasoning, enabling both answer generation and spatial localization. Fine-tuning MedGemma-4B-it on this dataset leads to a substantial improvement in VQA F1 (from 0.558 to 0.624, +11.8\%) while introducing spatial grounding capability (mean IoU = 0.615), without compromising text-based performance.

Importantly, 22.8\% of validation images achieve “good” localization (IoU $\geq$ 0.5), allowing visual verification of AI outputs—\textbf{which is} a critical step toward interpretable and clinically actionable VQA. Nonetheless, the model’s low recall (9.0\%) reflects the lack of multi-lesion training examples, whereas validation images average 8.3 lesions each. This highlights a key direction for improvement: incorporating lesion-dense training data to effectively bridge the distribution gap and enhance generalization.

Despite current limitations, the dataset’s balanced structure (41.7\% positive, 58.3\% negative) and clinically relevant question taxonomy provide a strong foundation for safe and explainable medical AI. Future work should enrich lesion diversity, explore structured supervision for multi-instance detection, and extend clinical validation across broader populations and institutions.

\section{Compliance with Ethical Standards}

This retrospective study utilized human subject data made publicly available through open access by the VinDr-CXR dataset~\cite{vindr-cxr}, provided by the Vingroup Big Data Institute. According to the license provided with the dataset, ethical approval was not required. The dataset is released under the CC BY 4.0 license.

\section{Acknowledgments}

This work was supported by the VinUni-Illinois Smart Health Center (VISHC). We thank the creators of the VinDr-CXR dataset for providing this valuable resource.

\bibliography{references}
\bibliographystyle{IEEEbib}
\end{document}